\documentclass[letterpaper, 10 pt, journal, twoside]{IEEEtran}

\usepackage{amsmath,amsfonts}
\usepackage{algorithmic}
\usepackage{algorithm}
\usepackage{array}
\usepackage[caption=false,font=normalsize,labelfont=sf,textfont=sf]{subfig}
\usepackage{textcomp}
\usepackage{stfloats}
\usepackage{url}
\usepackage{verbatim}
\usepackage{graphicx}
\usepackage{cite}
\usepackage{hyperref}
\usepackage{multirow}
\usepackage{bbding}
\usepackage{booktabs}
\usepackage{adjustbox}
\usepackage{color}
\usepackage{threeparttable}
\title{\fontsize{23.9pt}{28.68pt}\selectfont LXLv2: Enhanced LiDAR Excluded Lean 3D Object Detection with Fusion of 4D Radar and Camera}

\author{Weiyi Xiong$^1$, Zean Zou$^{2}$, Qiuchi Zhao$^1$, Fengchun He$^{2}$, and Bing Zhu$^{1,\dagger}$,~\IEEEmembership{Senior Member,~IEEE}%
\vspace{-5 mm}

\thanks{$^{1}$W.~Xiong, Q,~Zhao and B.~Zhu are with the School of Automation Science and Electrical Engineering, Beihang University, Beijing, P.R.~China.
{\tt\footnotesize \{weiyixiong, qiuchizhao, zhubing\}@buaa.edu.cn}}%
\thanks{$^{2}$Z.~Zou and F.~He are with the Continental Autonomous Mobility (Shanghai) Co., Ltd, Shanghai, P.R.~China. 
{\tt\footnotesize \{zean.zou, fengchun.he\}@continental-corporation.com}}%
\thanks{$^\dagger$Corresponding Author.}%
\thanks{This paper has been accepted by IEEE Robotics and Automation Letters. Digital Object Identifier 10.1109/LRA.2025.3536840}}

\begin{document}

\maketitle

\begin{abstract}
As the previous state-of-the-art 4D radar-camera fusion-based 3D object detection method, LXL utilizes the predicted image depth distribution maps and radar 3D occupancy grids to assist the sampling-based image view transformation. However, the depth prediction lacks accuracy and consistency, and the concatenation-based fusion in LXL impedes the model robustness. In this work, we propose LXLv2, where modifications are made to overcome the limitations and improve the performance. Specifically, considering the position error in radar measurements, we devise a one-to-many depth supervision strategy via radar points, where the radar cross section (RCS) value is further exploited to adjust the supervision area for object-level depth consistency. Additionally, a channel and spatial attention-based fusion module named CSAFusion is introduced to improve feature adaptiveness. Experimental results on the View-of-Delft and TJ4DRadSet datasets show that the proposed LXLv2 can outperform LXL in detection accuracy, inference speed and robustness, demonstrating the effectiveness of the model. 
\end{abstract}
\begin{IEEEkeywords}
4D imaging radar, camera, multi-modal fusion, 3D object detection, deep learning, autonomous driving.
\end{IEEEkeywords}

\vspace{-5mm}
\section{Introduction}\label{intro}

\IEEEPARstart{A}{utonomous} driving has become a hot topic in recent years. The whole process of autonomous driving can be divided into four stages: perception, prediction, planning and control\cite{UniAD}, among which perception is of great significance. To accurately perceive the environment, data from cameras \cite{LSS,BEVDepth}, LiDARs \cite{PointPillars,CenterPoint} and automotive radars \cite{rad_ins_seg} are commonly exploited.

As each sensor has its own pros and cons, researchers start to integrate information from multiple types of sensors. E.g., camera images are semantically rich but ambiguous in depth measuring, while LiDAR points are geometrically accurate but sparse \cite{BEVFusion}. Consequently, fusing camera and LiDAR modalities can make them complement each other, resulting in more accurate perception \cite{BEVFusion2, BiCo-Fusion}. However, the high cost of LiDARs hinders the widespread application of these methods. Although automotive radars are much cheaper and can work in all-weather conditions, traditional ones can only generate point clouds that are extremely sparse and noisy. In addition, they cannot measure the height information, which is necessary for 3D perception tasks such as 3D object detection \cite{VoD}. As a result, only a few works consider to perform camera-radar fusion-based 3D object detection \cite{CenterFusion,CRN}.

With the development of 4D radars, height information is accessible and the point cloud sparsity is reduced, so that several 4D radar-based \cite{RPFA-Net,SMURF} and 4D radar-camera fusion-based \cite{RCFusion} 3D object detection methods are proposed, obtaining promising results. 
As the previous state-of-the-art on both View-of-Delft\cite{VoD} and TJ4DRadSet\cite{TJ4DRadSet} datasets, LXL \cite{LXL} proposes to utilize image depth distribution maps and radar 3D occupancy grids to assist the sampling-based image view transformation (VT) \cite{Simple-BEV}. 

This paper focuses on two main limitations of LXL. First, although more accurate image VT can be performed with the assistance of radar 3D occupancy grids, it still struggles with imprecise image depth estimation. Moreover, the fusion module of LXL is simple concatenation and convolution, which is neither sufficient nor adaptive, making the detection accuracy decreases significantly when image quality drops. 

To enhance the depth estimation, some LiDAR-camera fusion-based methods \cite{BEVDepth} project LiDAR points onto image plane to generate sparse depth maps, serving as ground-truths of depth estimation. Extending this idea, in 4D radar-camera fusion-based approaches, it is intuitive to use radar points for substitution when LiDAR points are unavailable. Nevertheless, the sparsity and noise of radar points make the ground-truth depth map insufficient and inaccurate. Thus, we propose a one-to-many depth supervision strategy, where a radar point supervises an area of pixels centered at its projection. In addition, considering that depth remains consistent in a larger area for larger objects, we explicitly exploit the radar cross section (RCS) value, a unique characteristic of radar points reflecting the object size \cite{RCBEVDet}, to adjust the supervision area. Although LiDAR points can also be utilized for supervision in training process, we believe training without extra data is meaningful for online continual learning in the future.

As for the fusion module, there are three main methods to perform multi-modal fusion: concatenation-based \cite{LXL,Simple-BEV,RCBEV}, spatial/channel attention-based \cite{BEVFusion2,RCFusion,BiCo-Fusion} and cross attention-based \cite{CRN,RCBEVDet,DPFT}. 
Among them, spatial/channel attention-based fusion balances the speed and accuracy, but almost all models adopt only one of them. For example, BEVFusion\cite{BEVFusion2} applies a channel attention-based SE \cite{SE} fusion, and RCFusion \cite{RCFusion} utilzes a spatial attention-based IAM Fusion. In this work, we propose CSAFusion, which includes both channel attention and spatial attention for superior adaptiveness.  
According to predicted channel/spatial weights, the model can adaptively focus on the informative part of the feature map and filter out noise or irrelevant information. 

Contributions of this work can be summerized as follows:
\begin{itemize}
    \item Based on the previous state-of-the-art method LXL, we propose LXLv2, where enhancements of depth estimation and fusion methods are made to handle the limitations and improve the performance.
    
    \item To enhance image depth estimation, a one-to-many depth supervision strategy utilizing 4D radar points is introduced, exploiting geometric information from radar modality while alleviating the influence of inherent noise. In addition, an RCS-guided neighborhood determination method is devised to adjust the supervision area for different sizes of objects.

    \item To facilitate feature interaction among modalities and improve the model robustness, a channel and spatial attention-based fusion module named CSAFusion is introduced, such that the detection head can focus on informative channels and regions.

    \item Experiments on View-of-Delft/TJ4DRadSet show that our LXLv2 can outperform LXL by 1.8\%/1\% mAP$_\text{3D}$ with less inference time, and extensive ablation studies demonstrate the effectiveness of each component in LXLv2.
\end{itemize}

The rest of the paper are organized as follows. Related works, including camera-based, 4D radar-based, and 4D radar-camera fusion-based 3D object detection methods, are introduced briefly in Section \ref{related work}. 
The proposed enhancements of depth estimation and fusion module in LXLv2 are elaborated in Section \ref{method}. Experimental results and discussions are provided in Section \ref{experiments}. The conclusion is given in Section \ref{sec:conclusion}.

\vspace{-3mm}
\section{Related Work}\label{related work}
\subsection{Camera-based 3D Object Detection}
As 3D bounding boxes is usually predicted from features in the BEV/3D space, VT is needed in camera-based 3D object detection tasks. The VT methods can be mainly categorized into two types, i.e., depth-based VT \cite{BEVDepth,Simple-BEV,BEVDet,Dual-BEV,BEVSpread} and query-based VT \cite{BEVFormer,DETR3D,PETRv2}.

The depth-based VT explicitly utilizes the camera parameters and depth information to lift image pixels to frustum voxels or points, and further convert them to standard voxels or BEV maps. For instance, based on the pioneering work LSS\cite{LSS} in which the \textit{splatting}-based VT strategy is proposed, BEVDet\cite{BEVDet} applies the same method to perform 3D object detection. Differently, M$^2$BEV\cite{M2BEV} and Simple-BEV\cite{Simple-BEV} utilize the \textit{sampling} strategy under the assumption that image depths are uniformly distributed. To further improve the performance, Dual-BEV\cite{Dual-BEV} exploits both \textit{splatting} and \textit{sampling} strategies and fuses the resulted BEV features.

Intuitively, improved depth quality is beneficial for VT, so that some methods take the advantage of geometrically-accurate LiDAR points to introduce depth supervision. The earliest work BEVDepth\cite{BEVDepth} projects LiDAR points onto the image plane to generate a sparse depth map, which serves as the target of depth estimation. EA-LSS\cite{EA-LSS} further exploits the depth gradients in the ground-truth depth map, to realize edge-aware depth estimation.

In approaches applying the query-based VT like BEVFormer\cite{BEVFormer} and DETR3D\cite{DETR3D}, initialized 3D object queries are sent to transformer layers, to interact with image PV features through cross attention. After that, 3D bounding boxes are regressed from the updated object queries. Different from depth-based VT, query-based VT do not explicitly leverage image depths, bypassing the ill-posed 2D-3D projection. Nevertheless, the transformer itself introduces high complexity.

Although improvements has been witnessed recently, the performance of camera-based 3D object detection methods are still limited by the inherent depth ambiguity of images. 

\vspace{-3mm}
\subsection{4D Radar-based 3D Object Detection}

LiDAR-based methods dominate the field of 
3D object detection for years. With developments of 4D radars and the emerging 4D radar datasets \cite{VoD, TJ4DRadSet, K-Radar}, researchers start to study whether 4D radars can substitute LiDARs in 3D object detection. 
4D radar data can be represented as 3D points and other data formats such as 4D radar tensors. Except for few works \cite{K-Radar,RTNH+}, most methods take the point representation as input \cite{RPFA-Net,SMURF,RadarMFNet}, and modify models that are originally designed for LiDAR points. For example, based on PointPillars\cite{PointPillars}, RPFA-Net\cite{RPFA-Net} replaces the PointNet \cite{PointNet} by self-attention, while RadarPillarNet\cite{RCFusion} separate the feature extraction of position, Doppler and intensity in Pillar Feature Net. However, 4D radar point clouds are much sparser and noisier than LiDAR points, which is not explicitly handled in these work. To alleviate the influence of radar point sparsity, SMURF\cite{SMURF} add a kernel density estimation branch to extract sparsity-awareness features of radar points. 

Despite the increasing performance, these methods scarcely exploit the unique characteristics of radar points, such as the Doppler velocity and RCS value. 

\vspace{-3mm}
\subsection{4D Radar-Camera Fusion-based 3D Object Detection}

As 4D radar data contain geometric information and camera images have rich semantic information, it is intuitive to fuse these modalities and obtain higher detection accuracy. However, till now,  related methods are limited \cite{RCFusion,LXL,DPFT}.

RCFusion \cite{RCFusion} utilizes OFT\cite{OFT} to transform image features from PV to BEV, and proposes spatial-attention based IAM to fuse image and radar BEV features. To reduce the distortion of the image BEV features, LXL\cite{LXL} employs predicted image depth and radar 3D occupancy grids to assist the image VT, which significantly improves the performance. Differently, DPFT\cite{DPFT} takes 4D radar tensors as input and transform them to range-azimuth and azimuth-elevation maps. After that, a fusion and detection method similar to FUTR3D\cite{FUTR3D} is adopted.

In general, the recent 4D radar-camera fusion-based methods are comparable with LiDAR-based ones to a certain extent, showing the great potential of this field.
\begin{figure*}
    \centering
    \includegraphics[scale=0.5]{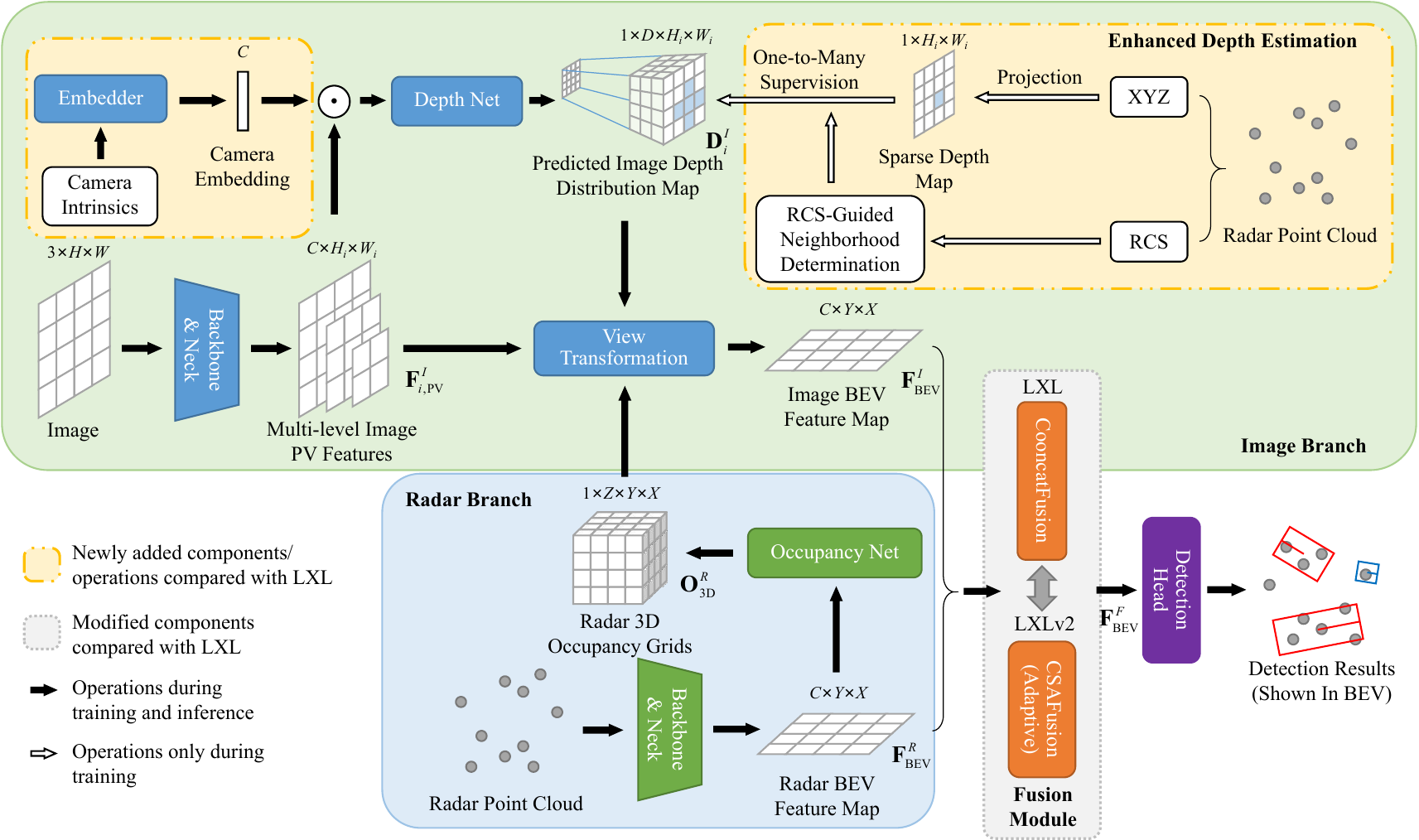} \vspace{-3mm}
    \caption{The overall architecture of LXLv2 compared with LXL \cite{LXL}. Differences lie in the depth estimation process and the fusion module. During depth estimation, camera intrinsics are introduced and radar points are exploited for one-to-many depth supervision, and RCS values are utilized to determine the supervision area. In the fusion module, CSAFusion is applied for improved feature adaptiveness and model robustness.}
    \label{fig:LXLv2} \vspace{-5mm}
\end{figure*}

\vspace{-3mm}
\section{A Brief Retrospection of LXL}
The overall framework of LXL\cite{LXL} is illustrated in Fig. \ref{fig:LXLv2}, and can be divided into four parts:

1) The \textbf{radar branch} extracts a radar BEV feature map $\mathbf F_\text{BEV}^R\in\mathbb R^{C\times Y\times X}$ and further predicts the radar 3D occupancy grids $\mathbf O_\text{3D}^R\in\mathbb R^{Z\times Y\times X}$ with an occupancy net:  \vspace{-1mm}
\begin{equation} 
    \mathbf O^R_\text{3D}=\mathtt{Sigmoid}(\mathtt{Conv}^{1\times1}_Z(\mathbf F^R_\text{BEV})),
\end{equation}\vspace{-1mm
}where $\mathtt{Conv}^{k\times k}_c$ denotes the convolution layer of kernal size $k\times k$ and output channel $c$, and $Z$ is the number of voxels in the height dimension.

2) The \textbf{camera branch} extracts image PV feature maps $\mathbf F_\text{PV}^I\in\mathbb R^{C\times H\times W}$, and estimates image depth distribution maps $\mathbf D^I$: \vspace{-1mm}
\begin{equation}
    \mathbf{D}^I_i=\mathtt{Softmax}(\mathtt{Conv}^{1\times1}_D(\mathbf{F}_{\text{PV},i}^I)),
\end{equation}\vspace{-1mm
}where the subscript $i$ represents the $i$-th feature level, and $D$ denotes the number of depth bins.

After that, $\mathbf F_\text{PV}^I$ is transformed to a BEV feature map $\mathbf F_\text{BEV}^I\in\mathbb R^{C\times Y\times X}$ with the assistance of $\mathbf O_\text{3D}^R$ and $\mathbf D^I$, utilizing the \textit{radar occupancy-assisted depth-based sampling} strategy:
\begin{equation}
\begin{aligned}
    \mathbf F_\text{3D}^I&=\mathtt{BilinearSample}(\mathtt{Proj}^{\text{2D}}_I(\mathbf V), \mathbf F_\text{PV}^I),\\
    \mathbf D_\text{3D}^I&=\mathtt{TrilinearSample}(\mathtt{Proj}^{\text{2.5D}}_I(\mathbf V), \mathbf D^I),\\
    \mathbf F_\text{BEV}^I&=\mathtt{Convs}^{3\times3}_{C,C,C}(\mathtt{Reshape}([\mathbf F_\text{3D}^I\odot \mathbf D_\text{3D}^I; \mathbf F_\text{3D}^I\odot \mathbf O_\text{3D}^R])),
\end{aligned}
\end{equation}
where $\odot$ is the element-wise production with broadcasting, and $[\cdot; \cdot]$ represents the channel concatenation. $\mathbf V\in\mathbb R^{3\times Z\times Y\times X}$ is the pre-defined voxel center coordinates, and $\mathtt{Proj}^{\text{2D}}_I(\mathbf V)$/$\mathtt{Proj}^{\text{2.5D}}_I(\mathbf V)$ means their 2D/2.5D projection on the image (i.e., $(u,v)$/$(u,v,d)$, where $(u,v)$ is the pixel coordinate and $d$ is the depth). The $\mathtt{Convs}^{k\times k}_{c_1,\cdots,c_n}$ denotes $n$ consecutive convolution layers with kernel size $k\times k$, and the number of output channels of the $j$-th layer is $c_j$. The $\mathtt{Reshape}(\cdot)$ operation transforms the shape of the input tensor from $(2C,Z,Y,X)$ to $(2C\cdot Z,Y,X)$.

In a word, the lifted image features $\mathbf F_\text{3D}^I$ are combined with the scene's geometry deduced from both camera and radar data. As image depth is ambiguous and radar occupancy is sparse, discarding either will result in sub-optimal VT.

3) The \textbf{fusion module} fuses $\mathbf F_\text{BEV}^R$ and $\mathbf F_\text{BEV}^I$ by concatenation and convolution: \vspace{-1mm}
\begin{equation}
    \mathbf F_\text{BEV}^F=\mathtt{Convs}^{3\times3}_{C,C}([\mathbf F_\text{BEV}^R; \mathbf F_\text{BEV}^I]).
\end{equation}  \vspace{-5mm}

4) Finally, the \textbf{detection head} predicts 3D bounding boxes based on the fused BEV featrue map $\mathbf F_\text{BEV}^F$.

\section{Proposed Method}\label{method}
To address the limitations of LXL mentioned in Section \ref{intro}, two modifications are made in LXLv2.

Specifically, to improve the accuracy of image depth estimation, camera parameters and radar points are exploited in the depth net. The former serve as a geometric prior for depth deducing, making the model to focus more on the geometry of the scene, and the latter provide course geometric information which is important to depth estimation. Considering that radar points are sparse and noisy, a one-to-many depth supervision strategy is proposed, making a radar point supervise multiple image pixels, and the position and area of the supervision region is determined by 3D coordinates and RCS value of the radar point, respectively. 

Furthermore, to adaptively mix the features from two modalities, an attention-based fusion module called CSAFusion is proposed, which is consist of a channel attention and a spatial attention. Concretely, the former focuses on the most informative channels, and the latter attends to the position where object might locate.

The yellow and gray parts in Fig.\ref{fig:LXLv2} represents the differences with LXL. Details of these enhancements are provided in the following subsections.

\vspace{-2mm}
\subsection{Enhancements of Image Depth Estimation}
As information lost is unavoidable during the 3D-to-2D projection, it is an ill-posed question to recover depth information from RGB images. LXL\cite{LXL} estimates the image depth distribution map with a convolution-based depth net, ignoring camera intrinsics and having no direct supervision. As a result, we introduce camera intrinsics and depth supervision to assist image depth estimation in LXLv2.

\textbf{Introducing Camera Intrinsics to Depth Estimation.} Following BEVDepth\cite{BEVDepth}, LXLv2 also utilize camera parameter embedding for depth estimation. 

Specifically, the inverse of camera intrinsic matrix $\mathbf K\in\mathbb R^{3\times 3}$ is embedded to the feature space using a linear layer and multiplied with image PV features before depth estimation. This is inspired by that $\mathbf K^{-1}$ is the pixel-to-camera coordinate transformation matrix, and depths are actually the $z$ coordinates of points in the camera coordinate system. Here, we do not take extrinsic parameters into account, as depth is unrelated to the image pose when predicted from the image space. In addition, only single-view image is available in our task and extrinsic parameters are the same across different scales of image PV features, so that intrinsics are enough for achieving cross-level depth consistency.

The embedding process can be formulated as
\begin{equation}
\begin{aligned}
    \mathbf E &=\mathtt{Linear}_{C}(\mathtt{Flatten}(\mathbf{K}_i^{-1}))\\
    \mathbf{D}^I_i&=\mathtt{Softmax}(\mathtt{Conv}^{1\times1}_D(\mathbf E\odot\mathbf{F}_{\text{PV},i}^I)),
\end{aligned}
\end{equation}
where $\mathbf E$ is the camera parameter embedding and $\mathtt{Linear}_c$ means a linear layer with output channel $c$. Note that the intrinsic matrix need to be modified to fit the resolution as feature maps are down-sampled. 

The introduction of camera intrinsics make the model easier to learn consistent depth across different feature scales. 

\textbf{One-to-Many Depth Supervision with 4D Radar Points.} Similar to BEVDepth\cite{BEVDepth}, radar points are projected onto the image plane to generate sparse depth map, which serves as the depth supervision. 

Considering that radar points are noisy and have relatively low angle resolution, misalignment are unavoidable during the projection process, so that a one-to-many depth supervision strategy is designed. Specifically, for each pixel $p$ that has ``ground-truth" depth $d^\text{gt}_p$ generated from radar points, we utilize $d^\text{gt}_p$ to supervise pixels in its neighborhood, and take the minimum loss as the final loss term:
\begin{equation}\label{eq:depth loss}
    L^\text{depth}_p(\mathbf D, d_p^\text{gt})=\min_{j\in\mathcal N(p)}\ell^\text{depth}(d_j, d_p^\text{gt}),
\end{equation}
where $\mathcal N(p)$ is the neighborhood of pixel $p$, and $\ell^\text{depth}(d_j, d_p^\text{gt})$ is the depth loss between the predicted depth distribution of pixel $j$ and the ground-truth depth of pixel $p$. 

Note that this one-to-many design can handle pixels that correspond to multiple radar points. Applying one-to-one supervision (i.e., $\mathcal N(p)=\{p\}$ in \eqref{eq:depth loss}) need to keep only one or calculate the average depth of these points, but in one-to-many supervision, each of these depths can be seen as a ground-truth depth and \eqref{eq:depth loss} can be calculated multiple times for pixel $p$.

The $\ell^\text{depth}$ in \eqref{eq:depth loss} contains two terms: the classification term and the regression term. The former takes the depth distribution prediction as a classification task, and regards the nearest depth bin $D_k$ to the ground-truth depth $d^\text{gt}_p$ as the ground-truth ``category". The latter calculate the expectation depth $\mathbb E(d_j)$ of the predicted depth distribution $d_j$:
\begin{equation}
    \mathbb E(d_j)=\sum_{l=1}^{N_D}p_{j,l}\bar d_l,
\end{equation}
where $\bar d_l$ is the medium depth of the $l$-th depth bin, and $p_{j,l}$ is the predicted probability of the $l$-th depth bin in $d_j$.

These two types of depth loss are both applied, because they focus on different aspects of improving the depth estimation performance. The classification loss aims to sharpen the depth distribution, but errors are introduced due to the discretization of ground-truth depths; the regression loss results in more accurate depth estimation, but the sharpness of the distribution cannot be guaranteed (e.g., if a ground-truth depth equals to the medium value in the depth range, a uniform distribution is enough to reduce the loss to zero).

In our work, cross entropy loss and L1 loss are applied as the classification loss and the regression loss, respectively. The final formulation of $\ell^\text{depth}$ is
\begin{equation}\label{eq:depth loss term}
    \begin{aligned}
    \ell^\text{depth}(d_j, d_p^\text{gt})=&\lambda_1\cdot\text{CrossEntropy}(d_j, D_k)\\
    +&\lambda_2\cdot\text{L1}(\mathbb E(d_j), d^\text{gt}_p),
\end{aligned}
\end{equation}
where $\lambda_1,\lambda_2$ are loss weights.

\textbf{RCS-Guided Neighborhood Determination.} It is important to properly define the neighborhood of a pixel $p$ that has ground-truth depth.

\begin{figure}
    \centering
    \includegraphics[scale=0.5]{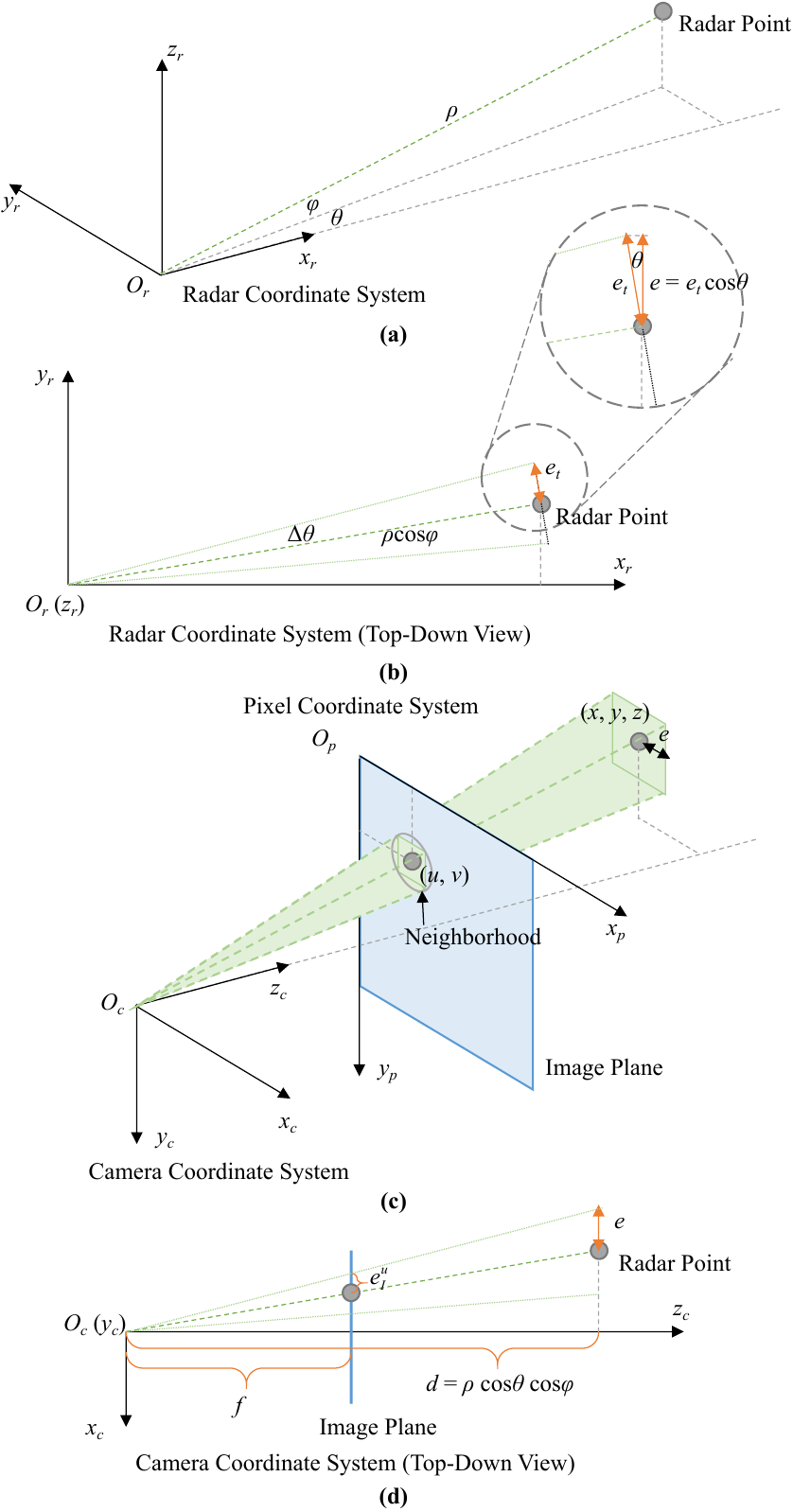}\vspace{-3mm}
    \caption{The illustration of maximum position error of radar points.}
    \label{fig:position error}\vspace{-5mm}
\end{figure}

The concept of neighborhood in loss calculation is introduced to address the misalignment in radar point projections caused by position errors. This raises a key question: what is the position error of radar points on the image plane? The maximum positional error of radar points increases linearly with range, while the projected length of a fixed line segment decreases linearly with increasing depth. Notably, it can be shown that all radar points exhibit approximately the same maximum positional error on the image plane.

As illustrated in Fig.\ref{fig:position error}(a), assume that there is a radar point $(\rho,\theta,\varphi)$ in radar sphere coordinate, where $\rho$, $\theta$ and $\varphi$ are range, azimuth and elevation angles, respectively. If the azimuth angle resolution of the radar sensor is $\Delta\theta$, the maximum tangential position error is $e_t=\rho\cos\varphi\cdot\Delta\theta$ (for brevity, ``maximum" is omitted in the following text), such that the error in $y_r$-axis is $e=e_t\cos\theta$, as shown in Fig.\ref{fig:position error}(b). 

To facilitate further analysis, we assume that the axes and origin of radar coordinate and camera coordinate are aligned, such that the horizontal position error of projected radar point on the image plane is $e_I^u=e\cdot f/d=f\cdot\Delta\theta$, where $f$ is the camera focal length, as shown in Fig.\ref{fig:position error}(c)(d). As $\theta$ is small in the camera field of view, $\Delta\theta$ can be regarded as a constant, so all radar points have approximately the same maximum horizontal position error $e_I^u=f\cdot\Delta\theta$ on the image plane. 

Similarly, it can be inferred that the maximum vertical positional error on the image plane $e_I^v=f\cdot\Delta\varphi$ is also irrelevant to the position of the point, where $\Delta\varphi$ is the elevation angle resolution. As a result, all radar points have approximately the same maximum positional error $e_I=f\cdot\sqrt{\Delta\theta^2+\Delta\varphi^2}$.

Based on the above derivation, the projection of all radar point can deviate by the same maximum distance. Consequently, the most straightforward way is to set the same neighborhood radius for all pixels to represent this deviation.

Additionally, we adjust the neighborhood radius according to the radar RCS value if provided. This design motivate from that RCS reflects the object size to some extent \cite{RCBEVDet}, and for larger objects, the depth consistency remains in a larger area. 

Specifically, given a point $\mathbf x=(x,y,z)$ in the camera coordinate system and the camera intrinsic matrix $\mathbf{K}$, we can project it onto the image plane according to
\begin{equation}\label{eq:cam2pix}
    \begin{bmatrix}
    u\\v\\1
    \end{bmatrix}\cdot d=\underbrace{\begin{bmatrix}
        f_x&0&c_x\\
        0&f_y&c_y\\
        0&0&1
    \end{bmatrix}}_{\mathbf K}\cdot\begin{bmatrix}
        x\\y\\z
    \end{bmatrix},
\end{equation}
where $(u,v)$ (denoted as $\mathbf p$ later) is the projected point coordinate in the pixel coordinate system, $d=z$ is the depth, $f_x,f_y$ are focal lengths, and $c_x,c_y$ are center offsets.

\begin{figure}
    \centering
    \includegraphics[scale=0.5]{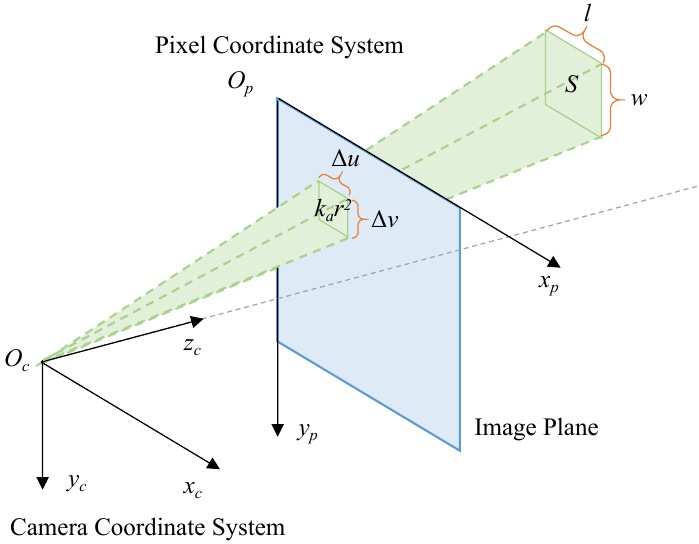}\vspace{-3mm}
    \caption{The illustration of neighborhood and object size.}
    \label{fig:rcs neighbor}\vspace{-5mm}
\end{figure}

Let $\mathbf x_1=[a+w,b+h,d]^T$ and $\mathbf x_2=[a,b,d]^T$ be two points in the camera coordinate system. According to \eqref{eq:cam2pix}, $\mathbf p_1$ and $\mathbf p_2$ can be obtained, and $\mathbf p_1-\mathbf p_2=[\Delta u,\Delta v]^T=[f_x\cdot w/d,f_y\cdot h/d]^T$. As illustrated in Fig.\ref{fig:rcs neighbor}, we let $\Delta u\cdot\Delta v\approx k_a\cdot r^2, w\cdot h\approx S$, where $k_a$ is the hyper-parameter, $r$ is the neighborhood radius, and $S$ is the object size. It can be obtained that
\begin{equation}
    r^2=\frac{f_x\cdot f_y\cdot S}{k_a\cdot d^2}.
\end{equation}

If the image is downsampled by a factor of $s(s>1)$, the focal lengths need to be divided by $s$, resulting in
\begin{equation}\label{eq:S to r2}
    r^2=\frac{f_x\cdot f_y\cdot S}{k_a\cdot d^2\cdot s^2}.
\end{equation}

With the assumption that the object size $S$ is proportional to RCS value (in m$^2$), i.e., $S=k_s\cdot v_\text{RCS}$, and considering that RCS values are usually in dBsm, we have
\begin{equation}\label{eq:rcs dB to S}
    S=k_s\cdot 10^{v_\text{RCS}^\text{dB}/10}.
\end{equation}

Based on \eqref{eq:S to r2} and \eqref{eq:rcs dB to S}, we can deduce that
\begin{equation}
    r=\frac{\sqrt{f_x\cdot f_y\cdot k_s/k_a}}{s\cdot d}\cdot 10^{v_\text{RCS}^\text{dB}/20}.
\end{equation}

As $k_s$ is unknown and $k_a$ is adjustable, $\sqrt{k_s/k_a}$ is substituted with $k$. Therefore, the neighborhood radius $r$ can be calculated by
\begin{equation}\label{eq:radius}
    r=\frac{k\cdot\sqrt{f_x\cdot f_y}}{s\cdot d}\cdot 10^{v_\text{RCS}^\text{dB}/20}.
\end{equation}

Finally, to avoid $r$ becoming too large, we clamp it by an upper threshold $r_{\max}$. 

Since RCS is possibly influenced by object material and incident angle, it is advantageous to decompose it into distinct components and utilize only the one pertaining to object size here. This will be included in our future work.

\vspace{-3mm}
\subsection{CSAFusion}
\begin{figure*}
    \centering
    \includegraphics[scale=0.53]{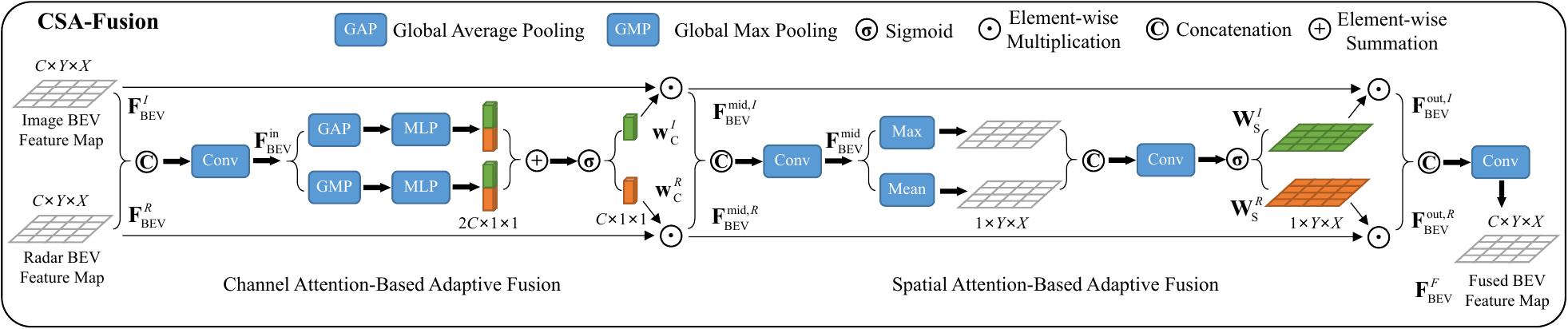}\vspace{-3mm}
    \caption{The architecture of CSAFusion.}
    \label{fig:CSAFusion} \vspace{-5mm}
\end{figure*}

To improve the model robustness and facilitate feature interaction, a channel and spatial attention-based fusion module is proposed in LXLv2. As illustrated in Fig.\ref{fig:CSAFusion}, it is composed of two parts: the channel attention-based adaptive fusion and the spatial attention-based adaptive fusion. 

Specifically, the channel attention-based adaptive fusion predicts channel weights for each modality to select the most informative channels, which can be formulated as
\begin{equation}
\begin{aligned}
    \mathbf F_\text{BEV}^\text{in}&=\mathtt{Conv}^{3\times3}_C([\mathbf F_\text{BEV}^R;\mathbf F_\text{BEV}^I]),\\
    \mathbf w_\text{C}^{X}&=\mathtt{Sigmoid}(\mathtt{MLP}_{C/3,C}(\mathtt{GAP}(\mathbf F_\text{BEV}^\text{in}))\\
    &~~~~~~~~~~~~+\mathtt{MLP}_{C/3,C}(\mathtt{GMP}(\mathbf F_\text{BEV}^\text{in}))),\\
    \mathbf F_\text{BEV}^{\text{mid},X}&=\mathbf w_\text{C}^{X}\odot\mathbf F_\text{BEV}^X,
\end{aligned}
\end{equation}
where $X\in\{R,I\}$ represents one of the modalities, $\mathbf w_\text{C}^{X}$ is the channel weights with respect to modality $X$, and $\mathbf F_\text{BEV}^{\text{mid},X}$ is the channel-attentive BEV features. $\mathtt{MLP}_{c_1,\cdots,c_n}$ denotes an MLP with $n$ linear layers, and the number of output channels of the $j$-th layer is $c_j$. $\mathtt{GAP}$ and $\mathtt{GMP}$ are the global average pooling and global max-pooling, respectively. The weight prediction formula follows CBAM\cite{CBAM}, but unlike RCFusion\cite{RCFusion} which utilize single-modal features $\mathbf F_\text{BEV}^X$ to predict the attention weights $\mathbf w^X$, the mixed feature $\mathbf F_\text{BEV}^\text{in}$ is adopted in CSAFusion because the contained information is more comprehensive.

Subsequently, the spatial attention-based adaptive fusion outputs spatial weights for both modalities, to filter out background noises, ignore irrelevant areas, and highlight the regions that have high probability of containing an object:
\begin{equation*}
\begin{aligned}
    \mathbf F_\text{BEV}^\text{mid}&=\mathtt{Conv}^{3\times3}_C([\mathbf F_\text{BEV}^{\text{mid},R};\mathbf F_\text{BEV}^{\text{mid},I}]),\\
    \mathbf W_\text{S}^{X}&=\mathtt{Sigmoid}(\mathtt{Conv}^{7\times7}_C([\mathtt{Max}(\mathbf F_\text{BEV}^\text{mid});\mathtt{Mean}(\mathbf F_\text{BEV}^\text{mid})])),\\
    \mathbf F_\text{BEV}^{\text{out},X}&=\mathbf W_\text{S}^{X}\odot\mathbf F_\text{BEV}^{\text{mid},X},\\
    \mathbf F_\text{BEV}^F&=\mathtt{Conv}^{3\times3}_C([\mathbf F_\text{BEV}^{\text{out},R};\mathbf F_\text{BEV}^{\text{out},I}]),
\end{aligned}
\end{equation*}
where $\mathtt{Max}$ and $\mathtt{Mean}$ are operated along the channel dimension. Similar to channel attention, the attention weights $\mathbf W_\text{S}^X$ are also predicted from the mixed feature $\mathbf F_\text{BEV}^\text{mid}$.

Finally, the fused BEV feature map $\mathbf F_\text{BEV}^F$, which is both channel-attentive and spatial-attentive, will be sent to the detection head for 3D bounding box prediction.

\vspace{-2mm}
\section{Experiments}\label{experiments}
\subsection{Datasets and Evaluation metrics}
We follow LXL\cite{LXL} to conduct experiments on two publicly-available datasets: View-of-Delft (VoD)\cite{VoD} and TJ4DRadSet\cite{TJ4DRadSet}. Both datasets contain RGB images from a monocular camera and point clouds from a 4D radar, and are annotated with 3D bounding boxes.

The VoD dataset is collected in Delft, Netherlands, and three categories of objects are mainly considered: cars, pedestrians and cyclists. The recordings are carefully selected by the creators to form a class-balanced dataset.

The TJ4DRadSet is recorded in Suzhou, China, and contains data under complicated road conditions or in dark/over-illuminated environments. Cars, pedestrians, cyclists and trucks are annotated in this dataset, where car is the dominant category. Therefore, the detection on TJ4DRadSet is more challenging than that on VoD.

To evaluate model performances, we apply mAP as evaluation metric for both datasets. Specifically, 3D APs of the entire annotated area (AP$_\text{3D}$) is calculated for evaluations on VoD. For TJ4DRadset, 3D/BEV APs within a range of 70 meters are reported. Please refer to \cite{LXL} for more detailed information.

\begin{table}\vspace{-5mm}
\centering
\scriptsize
\caption{Comparison with state-of-the-arts on the VoD \cite{VoD} \texttt{val} set.}\vspace{-1mm}\label{tab:vod results}
\begin{threeparttable}[b]
\begin{tabular}{cc|ccc|c|c}
\toprule
\multirow{2}{*}{Model} & \hspace{-1.5em}\multirow{2}{*}{Modality}\hspace{-0.5em} & \multicolumn{4}{c|}{AP$_\text{3D}$ (\%)} & \multirow{2}{*}{FPS\tnote{*}}\\ \cmidrule{3-6} 
 &  & Car\hspace{-0.8em} & Ped.\hspace{-0.8em} & Cyc. & mAP \\ \midrule
CenterPoint (CVPR'21) \cite{CenterPoint} & \hspace{-1em}R & 32.75\hspace{-0.8em} & 39.65\hspace{-0.8em} & 68.13 & 46.84 & 44.7\\
SMURF (T-IV'24) \cite{SMURF} & \hspace{-1em}R & 42.31\hspace{-0.8em} & 39.09\hspace{-0.8em} & 71.50 & 50.97 & 30.0\\  
FUTR3D (CVPR'23) \cite{FUTR3D} & \hspace{-1em}R+C & 46.01\hspace{-0.8em}  & 35.11\hspace{-0.8em} & 65.98 &  49.03 & 7.3\\
BEVFusion (ICRA'23) \cite{BEVFusion} & \hspace{-1em}R+C &  37.85\hspace{-0.8em} & 40.96\hspace{-0.8em} & 68.95 & 49.25 & 7.1\\
RCFusion (T-IM'23) \cite{RCFusion} & \hspace{-1em}R+C & 41.70\hspace{-0.8em} & 38.95\hspace{-0.8em} & 68.31 & 49.65 & - \\
RCBEVDet (CVPR'24) \cite{RCBEVDet} & \hspace{-1em}R+C & 40.63\hspace{-0.8em} & 38.86\hspace{-0.8em} & 70.48 & 49.99 & - \\
LXL (T-IV'24) \cite{LXL} & \hspace{-1em}R+C & 42.33\hspace{-0.8em} & \textbf{49.48}\hspace{-0.8em} & 77.12 & 56.31 & 6.1\\ \midrule
LXLv2 (\textbf{Ours}) & \hspace{-1em}R+C & \textbf{47.81}\hspace{-0.8em} & 49.30\hspace{-0.8em} & \textbf{77.15} & \textbf{58.09} & 6.5\\ \bottomrule
\end{tabular}
\begin{tablenotes}
    \item[*] The inference speed is evaluated on a single NVIDIA Tesla V100 GPU. \vspace{-1mm}
\end{tablenotes}
\end{threeparttable}
\end{table}

\vspace{-3mm}
\subsection{Implementation Details}
In the following experiments, the depth loss weights $\lambda_1,\lambda_2$ in \eqref{eq:depth loss term} are both set to $0.1$. For VoD dataset, the $k$ in \eqref{eq:radius} is set to $0.1$, and the maximum radius $r_{\max}$ is set to $2$. For TJ4DRadSet, we fix $r$ to 2, as RCS values are not provided. The other hyper-parameter settings are kept the same with LXL \cite{LXL}, except for decreasing the number of channels $C$ to 256 after the backbone and neck to alleviate over-fitting.

The training details is in consistent with LXL, including the optimizer, learning rate scheduler, image backbone pre-training and data/feature augmentation strategy. 

\subsection{Results and Analysis}

\begin{figure}
    \centering
    \begin{tabular}{@{}c@{}c@{}}
        \rotatebox[origin=c]{90}{\scriptsize ~~~~~LXLv2(\textbf{Ours})~~~~~~~~~~~~~~~~~~
LXL (T-IV'24) \cite{LXL}~~~~~~~~~~~~~~~Image} & \raisebox{-.5\height}{\includegraphics[scale=0.9]{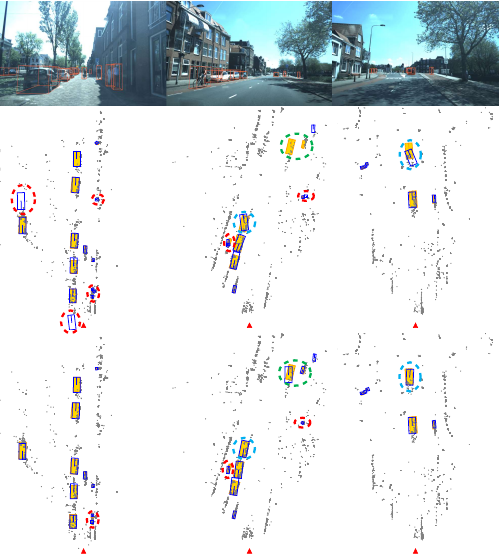}}
    \end{tabular}\vspace{-1mm}
    \caption{Visualization results of LXL\cite{LXL} and LXLv2 on the VoD \cite{VoD} \texttt{val} set (best viewed in zoom and color). Each column corresponds to a frame of data, containing an image and radar points (gray points) in BEV, where orange boxes represent ground-truths and blue boxes stand for predicted bounding boxes. The red triangle denotes the position of the ego-vehicle.}
    \label{fig:vod vis}\vspace{-3mm}
\end{figure}

\textbf{Results on VoD.} Table \ref{tab:vod results} reports experimental results on VoD validation set. LXLv2 ranks the first among all radar-based and radar-camera fusion-based methods, and outperforms the state-of-the-art LXL\cite{LXL} by 1.8\% mAP$_\text{3D}$. Specifically, the detection accuracy of LXLv2 on car category increases 5.5\%, as large objects typically correspond to more radar points, resulting in more accurate depth estimation. In contrast, small objects occupy a small area in BEV, which is more susceptible to surrounding background noise when predicting the spacial attention weights, so that the performance gain is limited.

In aspect of inference speed, LXLv2 surpasses LXL by 7\% due to decreased channel number, resulting in more balanced speed and accuracy. Although still slower than BEVFusion \cite{BEVFusion} by 13ms, the performance of LXLv2 is nearly 9\% higher. 

Typical visualization results are shown in Fig.\ref{fig:vod vis}. It can be seen that compared to LXL, LXLv2 can detect some distant or occluded objects more accurately.

\textbf{Results on TJ4DRadSet.} The results of different models on the test set of TJ4DRadSet are exhibited in Table \ref{tab:tj results} to show the generalizability of LXLv2. Although we cannot exploit the RCS value in TJ4DRadSet, the performance of LXLv2 still improves 1\% mAP$_\text{3D}$ and 1.2\% mAP$_\text{BEV}$ with fixed neighborhood size in depth supervision. 

\begin{table}[t]
\caption{Comparison with state-of-the-arts on TJ4DRadSet \cite{TJ4DRadSet} \texttt{test} set.}\vspace{-1mm}
\label{tab:tj results}
\centering
\begin{tabular}{cc|cc}
\toprule
Model & Modality & mAP$_\text{3D}$(\%) & mAP$_\text{BEV}$(\%) \\ \midrule
CenterPoint (CVPR'21) \cite{CenterPoint} & R & 30.79 & 38.42 \\
SMURF (T-IV'24) \cite{SMURF} & R & 32.99 & 40.98 \\
FUTR3D (CVPR'23) \cite{FUTR3D} & R+C & 32.42 & 37.51 \\
BEVFusion (ICRA'23) \cite{BEVFusion} & R+C & 32.71 & 41.12 \\
RCFusion (T-IM'23) \cite{RCFusion} & R+C & 33.85 & 39.76 \\
LXL (T-IV'24) \cite{LXL} & R+C & 36.32 & 41.20 \\
\midrule
LXLv2 (\textbf{Ours}) & R+C & \textbf{37.32} & \textbf{42.35} \\ \bottomrule
\end{tabular}\vspace{-3mm}
\end{table}

To demonstrate LXLv2's robustness under varying lighting conditions, Table \ref{tab:lighting breakdown} presents a performance breakdown. Note that CenterPoint \cite{CenterPoint} is the radar-only version of LXL \cite{LXL}, and all models in the table share the same channel numbers for fair comparison. Unlike LXL, whose performance drops in over-illuminated (shiny) conditions after fusing with the camera modality, LXLv2 outperforms its single-modality counterpart across all scenarios. This improvement is due to the CSAFusion module, which prioritizes radar modality when image quality declines, instead of treating both modalities equally.

\begin{table}[t]
\centering
\scriptsize
\caption{Performances of our model on different lighting conditions.} \vspace{-1mm}\label{tab:lighting breakdown}
\begin{tabular}{c|ccc|ccc}
\toprule
\multirow{2}{*}{Model} & \multicolumn{3}{c|}{mAP$_\text{3D}$(\%)} & \multicolumn{3}{c}{mAP$_\text{BEV}$(\%)} \\ \cmidrule{2-7} 
 & Dark & Standard & Shiny & Dark & Standard & Shiny \\ \midrule
CenterPoint\cite{CenterPoint} & 16.45 & 29.29 & 21.68 & 21.17 & 36.99 & 29.98 \\
LXL\cite{LXL} & 20.10 & 42.61 & 20.63 & 25.08 & 46.01 & 26.94 \\ 
LXLv2 (\textbf{Ours}) & \textbf{22.38} & \textbf{45.70} & \textbf{22.70} & \textbf{25.65} & \textbf{49.79} & \textbf{30.38} \\ \bottomrule
\end{tabular}\vspace{-5mm}
\end{table}

\begin{table}[]
\scriptsize
\centering \vspace{-5mm}
\caption{Ablation studies of depth estimation enhancements on VoD \cite{VoD} \texttt{val} set.} \vspace{-1mm}\label{tab:depth ablation}
\begin{tabular}{c|cccc|c}
\toprule
& Cam. Param. Emb. & Depth Sup. & \hspace{-1em}Neighborhood & \hspace{-1em}Loss & mAP$_\text{3D}$(\%) \\ \midrule
(a) & \XSolidBrush & \XSolidBrush &  &  & 56.26 \\
(b) & Intrinsics & \XSolidBrush &  &  & 57.04 \\
(c) & Intrinsics & One-to-One & \hspace{-1em}\XSolidBrush &  & 56.63 \\
(d) & Intrinsics &  One-to-Many & \hspace{-1em}Fixed & \hspace{-1em}min & 57.20 \\
(e) & Intrinsics & One-to-Many & \hspace{-1em}Dynamic & \hspace{-1em}min & \textbf{58.09} \\ \midrule
(f) & \hspace{-0.5em}Intrinsics \& Extrinsics\hspace{-0.5em} & One-to-Many & \hspace{-1em}Dynamic & \hspace{-1em}min & 57.29  \\
(g) & Intrinsics & One-to-Many & \hspace{-1em}Dynamic & \hspace{-1em}max & 57.40  \\
\bottomrule
\end{tabular}
\end{table}

\begin{table}[]
\centering 
\caption{Ablation studies of fusion module on the VoD \cite{VoD} \texttt{val} set.}\vspace{-1mm} \label{tab:fusion ablation}
\begin{tabular}{cc|c}
\toprule
Fusion Module & Attention & mAP$_\text{3D}$(\%) \\ \midrule
Concatenation (in LXL \cite{LXL}) & - & 55.81 \\
SE (in BEVFusion \cite{BEVFusion2}) & Channel &  56.37 \\ 
AWF (in BiCo-Fusion \cite{BiCo-Fusion}) & Spatial & 57.23 \\ 
IAM (in RCFusion \cite{RCFusion}) & Spatial & 57.77  \\ 
AGF (in LiRaFusion \cite{LiRaFusion}) & Channel \& Spatial & 56.19 \\ \midrule
CSAFusion (\textbf{Ours}) & Channel \& Spatial & \textbf{58.09} \\
\bottomrule
\end{tabular}\vspace{-5mm}
\end{table}

\vspace{-3mm}
\subsection{Ablation Study}
\textbf{Ablation on depth estimation enhancements.} 
To evaluate the impact of key designs in the proposed depth estimation enhancements, models with different depth supervision (Depth Sup.) configurations were tested, with results shown in Table \ref{tab:depth ablation}. Adding camera parameter embedding (Cam. Param. Emb.) to the baseline (a) improves the mAP of (b) by 0.8\%, enabling more consistent depth prediction across feature levels. Applying one-to-one depth supervision, similar to LiDAR-supervised approaches \cite{BEVDepth}, reduces detection performance in experiment (c) due to radar measurement inaccuracies and potential projection misalignment. In contrast, adopting the proposed one-to-many supervision with a fixed neighborhood radius increases the mAP of (d) by 0.6\%, surpassing (b) and highlighting the importance of accounting for radar point position errors. Finally, dynamically adjusting the neighborhood radius based on RCS values further improves performance by 0.9\% in experiment (e), confirming the effectiveness of the proposed enhancements.

To prove the effectiveness of our camera parameter embedding and loss design, additional experiments are conducted.
After including camera extrinsic parameters in the embedding, the performance of (f) drops 0.8\%, as mixing with irrelevant information can impede learning.
In experiment (g), the $\min$ operation in \eqref{eq:depth loss} is changed to $\max$, resulting in a 0.7\% mAP decline. This is because $\max$ forces the target depth of all neighboring pixels to be the point depth, which is not always true, as foreground pixels and background pixels may be adjacent on the image plane.

\textbf{Ablation on the fusion module.} In Table \ref{tab:fusion ablation}, the proposed CSAFuson is compared with fusion methods applied in other models, including the original concatenation-based fusion in LXL\cite{LXL}, SE-based fusion in BEVFusion \cite{BEVFusion2}, adaptive weighted fusion (AWF) in BiCo-Fusion \cite{BiCo-Fusion}, IAM in RCFusion\cite{RCFusion} and adaptive gated fusion (AGF) in LiRaFusion \cite{LiRaFusion}. Note that AGF can be regarded as a channel-spatial joint attention because it simultaneously predicts a weight for each channel at each location of the input feature map. The results show that the concatenation-based fusion performs worst, while spatial attention-based fusion performs superior than channel attention-based fusion. After combining channel attention and spatial attention, our CSAFusion yields the optimal performance, but AGF fails as the entanglement of channel and space brings difficulty and complexity for learning on a relatively small dataset.

\vspace{-3mm}
\section{Conclusion}\label{sec:conclusion}
In this paper, we propose LXLv2, a 4D radar-camera fusion-based 3D object detection method. In LXLv2, two enhancements are made to overcome limitations of LXL, including one-to-many depth supervision strategy with RCS-guided neighborhood determination, and CSAFusion module. The former exploits geometric information in the radar modality to enhance image depth estimation, while the latter utilizes the channel attention and the spatial attention to improve performance and robustness. Experiments show that the proposed LXLv2 outperforms the state-of-the-art LXL on both VoD and TJ4DRadSet datasets with higher inference speed. 

\vspace{-3mm}
\bibliographystyle{IEEEtran}

\bibliography{reference}

\end{document}